\documentclass{article}
\usepackage{spconf,amsmath,epsfig}
\usepackage{xurl}

\let\OLDthebibliography\thebibliography
 \renewcommand\thebibliography[1]{
   \OLDthebibliography{#1}
   \setlength{\parskip}{0pt}
   \setlength{\itemsep}{0pt plus 0.3ex}
 }

\title{EOD: The IEEE GRSS Earth Observation Database}
\name{Michael Schmitt\textsuperscript{1}, Pedram Ghamisi\textsuperscript{2,3}, Naoto Yokoya\textsuperscript{4,5}, Ronny H\"ansch\textsuperscript{6}}
\address{\textsuperscript{1}Department of Aerospace Engineering, University of the Bundeswehr Munich, Germany\\
\textsuperscript{2}Helmholtz-Zentrum Dresden-Rossendorf, Germany\\
\textsuperscript{3}Institute of Advanced Research in Artificial Intelligence, Austria\\
\textsuperscript{4}The University of Tokyo, Japan\\
\textsuperscript{5}RIKEN Center for Advanced Intelligence Project, Japan\\
\textsuperscript{6}German Aerospace Center (DLR), Microwaves and Radar Institute, SAR Technology}
\begin{document}
%
\maketitle
\begin{abstract}
In the era of deep learning, annotated datasets have become a crucial asset to the remote sensing community. In the last decade, a plethora of different datasets was published, each designed for a specific data type and with a specific task or application in mind. In the jungle of remote sensing datasets, it can be hard to keep track of what is available already. With this paper, we introduce EOD - the IEEE GRSS Earth Observation Database (EOD) - an interactive online platform for cataloguing different types of datasets leveraging remote sensing imagery.
\end{abstract}
\section{Introduction}
\label{sec:intro}

Annotated datasets for the purpose of training and evaluating machine learning models have become a vital requirement for many branches of cutting-edge research in remote sensing and Earth observation. In contrast to general computer vision, whose target is the extraction of information from every-day images containing every-day objects, such as furniture, animals, or road signs, there is a much greater variety of image modalities and tasks in remote sensing. In this field, people may want to work with classical optical imagery, make use of multi- or hyperspectral sensing capabilities, or resort to even more special sensor technologies such as laser scanning or synthetic aperture radar. 
In addition to this there is a difference in scale with spatial resolutions ranging from centimeters to kilometers and temporal resolutions ranging from minutes (e.g. with drone-based imagery) to several weeks. 
Moreover, there is a multitude of applications including object detection and recognition (e.g. ship and oil spill detection), semantic segmentation (e.g. land cover mapping), instance segmentation (e.g. building footprint extraction), but also regression tasks such as the estimation of bio-/geo-physical parameters (e.g. building height, soil moisture, biomass, etc.).
Given this diversity it becomes clear that we will not see a single extremely large all-purpose image database, such as ImageNet\footnote{As a prime example for an annotated computer vision dataset, ImageNet contains more than 14 million images depicting objects from more than 20,000 categories.} in the near future. Thus, as e.g. discussed by \cite{Schmitt2021}, we are overwhelmed by newly published remote sensing datasets, all of which are annotated with a specific combination of sensor modality, application, and geographic location. In order to structure the available datasets and to provide an overview, the Image Analysis and Data Fusion (IADF) Technical Committee of IEEE GRSS has created the Earth Observation Database (EOD), which is described in this paper. 

\begin{figure}[ht!]
    \centering
    \includegraphics[width=\linewidth]{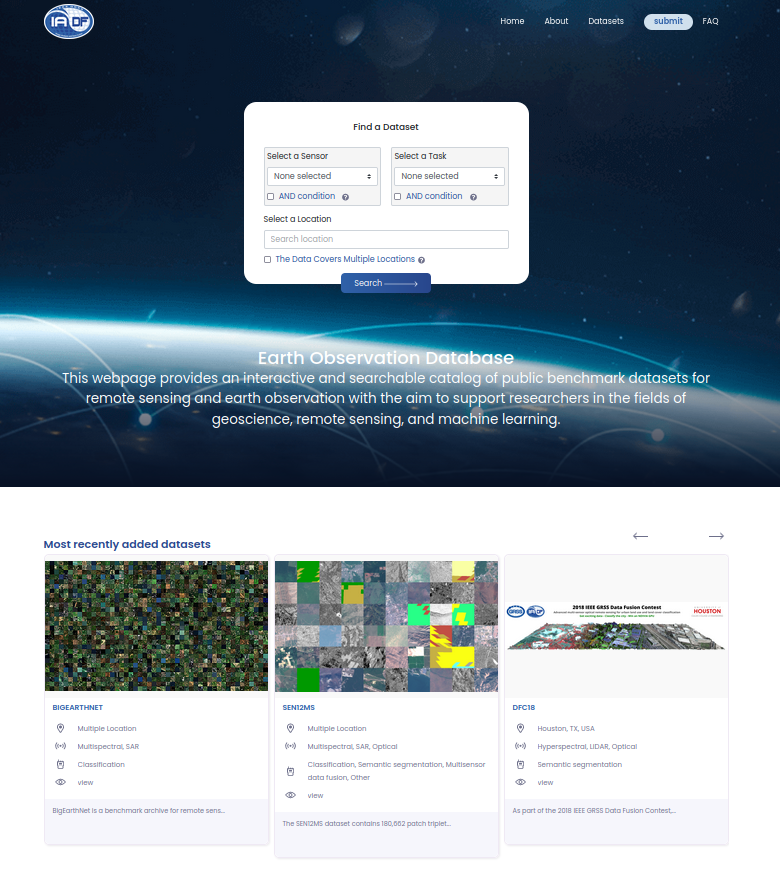}
    \caption{The landing page of the EOD website.}
    \label{fig:EOD_LandingPage}
\end{figure}

\section{Related Work}
To the best of the authors' knowledge there is not yet one central database with the goal to provide a comprehensive catalogue of existing remote sensing datasets. 
Data itself can be accessed via web portals of the corresponding institutions that collected the data, e.g. the Copernicus data hub \cite{copernicus}, or open data programs of companies (e.g. \cite{maxar,capella}). 
While this might include data such as semantic maps that can serve as reference for training and evaluation (e.g. the CORINE land cover \cite{corine}), these databases are limited to very specific sensors and not intended to serve as a general purpose catalogue for Earth observation datasets. 
Benchmark datasets are often only listed on the web pages of individual research labs or institutions, which summarize all datasets they have published, e.g. \cite{xxz}, \cite{whu} or \cite{isprs}. 
Competition sites such as Kaggle\footnote{\url{https://www.kaggle.com/datasets}} or CodaLab\footnote{\url{https://codalab.lisn.upsaclay.fr/}} as well as data storage sites such as IEEE DataPort\footnote{\url{https://ieee-dataport.org/}}  contain many datasets of relevance but have a very different focus. 
They contain data from all kinds of scientific disciplines, the provided information is not standardized, and the search capabilities are not tailored towards factors of importance in Earth observation and remote sensing. 
Evaluation servers such as DASE\footnote{\url{http://dase.grss-ieee.org/}} provide access to public leaderboards and allow a fair and unbiased evaluation based on standardized protocols. However, the number of datasets is usually limited. 
Cloud services such as Amazon Web Services\footnote{\url{https://registry.opendata.aws/}}, Microsoft's Planetary Computer\footnote{\url{https://planetarycomputer.microsoft.com/catalog}}, Radiant Earth's MLHub\footnote{\url{https://www.mlhub.earth/}}, etc. provide access to selected datasets. 
While they do aim to have a comprehensive list of datasets available, they are limited to those that have been added into their databases. 
What all of these commendable efforts have in common is that they focus on specific aspects such as data hosting, evaluation, easy access for processing in the cloud, etc., that are different from simply cataloguing existing datasets. 
This leads to databases that either contain a huge set of very diverse datasets (potentially including but not limited to Earth observation data) in a rather unstructured way, or consist of only a few carefully selected datasets. 

What is missing so far is a central catalogue that provides an exhaustive list of available datasets with their basic information, can be accessed and extended by the community, and queried in a structured and interactive manner. 
One of the most extensive and best-curated lists is the \emph{Awesome Satellite Imagery Datasets} list \cite{asid}. While it is an open github project welcoming contributions by anybody interested, it's not a database with query functionality. 
EOD aims to close this gap by providing an interactive catalogue, curated by GRSS IADF, extendable and accessible by everybody, as an easy access point to search for Earth observation datasets.

\section{The Earth Observation Database}
\label{sec:EOD}
The EOD web page is reachable via \url{https://eod-grss-ieee.com/}. Its landing page is depicted in Fig.~\ref{fig:EOD_LandingPage}. It displays both a query form for immediate search on available datasets and advertises those datasets that were either most recently added to the database or most popular. From this starting point, three basic options are possible:
\begin{itemize}
    \item Starting an immediate search for a datasets following specified requirements using the landing page query form.
    \item Navigating to the full data catalogue by clicking on the \emph{Datasets} link in the top frame of the landing page.
    \item Submitting a new dataset by clicking on the \emph{Submit} link in the top frame of the landing page.
\end{itemize}
Those three options are described in detail in the following.

\subsection{Dataset Query}
\begin{figure}[ht!]
    \centering
    \includegraphics[width=\linewidth]{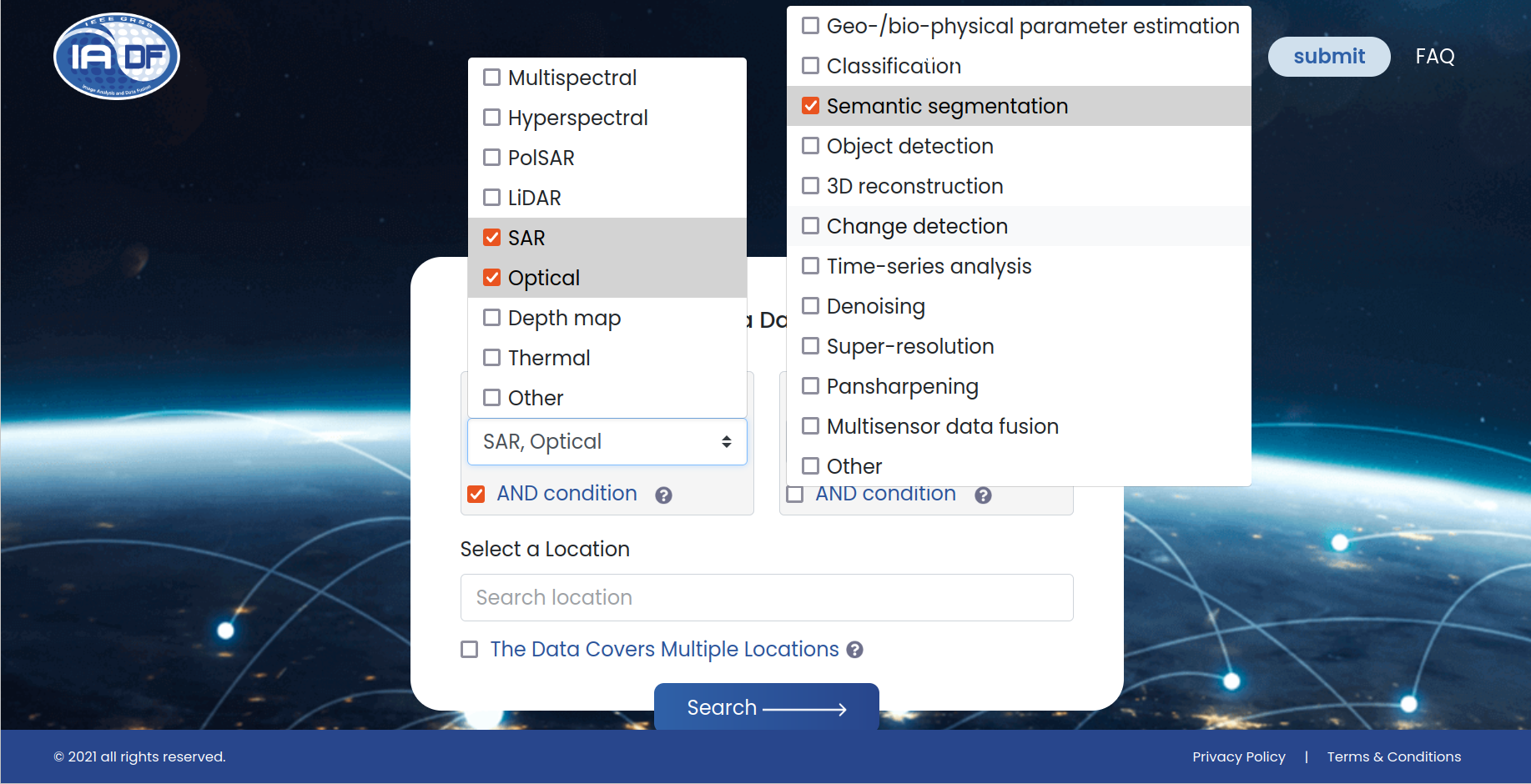}
    \caption{The EOD query interface.}
    \label{fig:EOD_Query}
\end{figure}

The simple query interface including the current selection options is shown in Fig.~\ref{fig:EOD_Query}. Generally, there are three features that can be used to identify relevant datasets:
\begin{itemize}
    \item The underlying sensor technology.
    \item The desired task/application.
    \item The location of the dataset.
\end{itemize}
As an example, the query shown in Fig.~\ref{fig:EOD_Query} would search the database for datasets containing both SAR and optical data annotated for semantic segmentation purposes. 
Note that the selection of multiple options is possible. A checkbox controls whether they are combined by a logical \emph{AND} or \emph{OR}. 
The location field is still left open, though. Note that either data from a specific location or datasets covering multiple locations can be requested.

\begin{figure*}[ht!]
    \centering
    \includegraphics[width=0.8\linewidth]{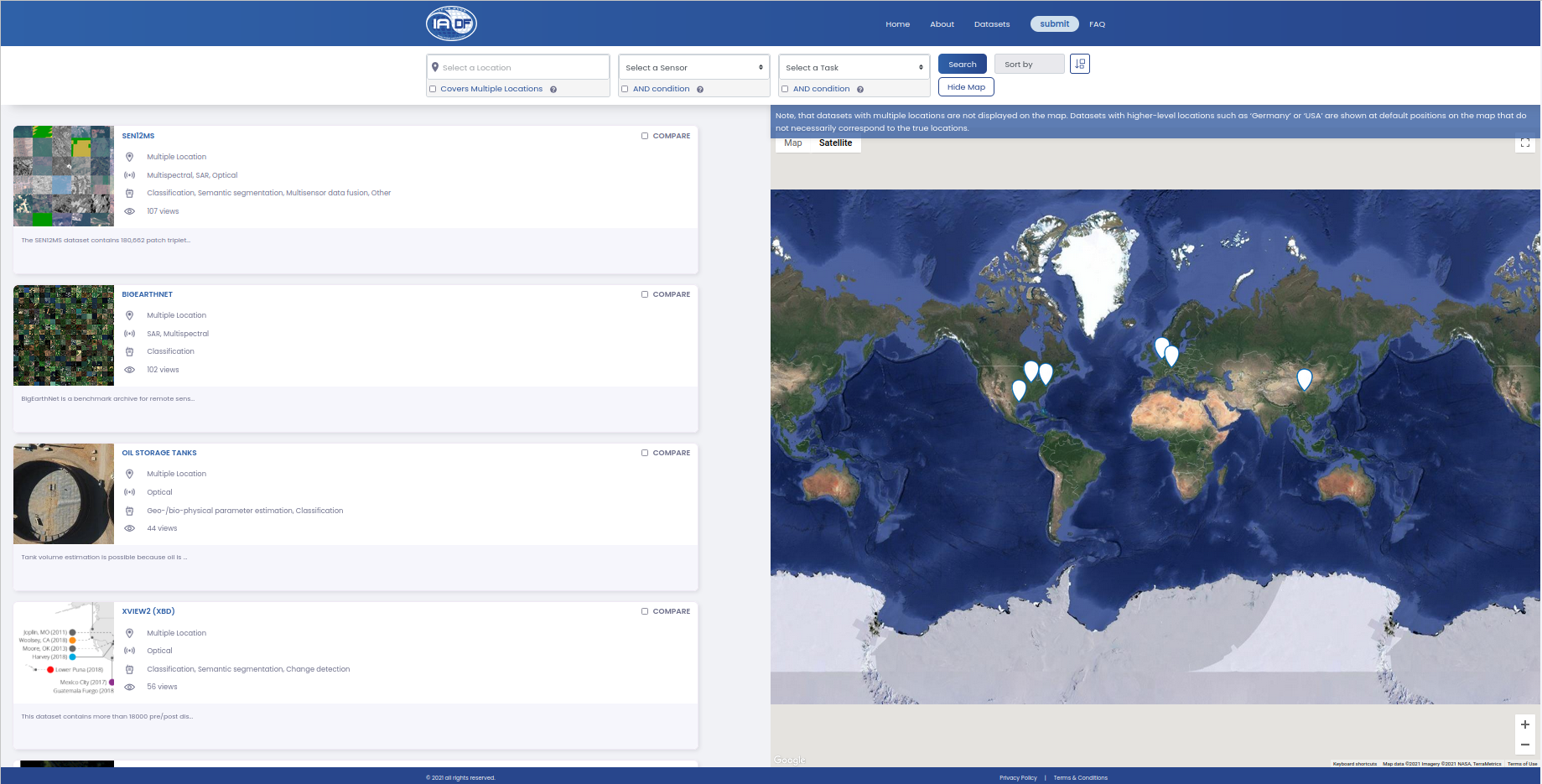}
    \caption{The map view of the EOD data catalogue.}
    \label{fig:EOD_Map}
\end{figure*}

\subsection{Data Catalogue}\label{sec:catalogue}
The core of the EOD is the database of user-submitted datasets. It can be viewed either in an illustrated list view, or in an interactive map view indicating those datasets that cover only a specific location with a marker (see Fig.~\ref{fig:EOD_Map}). 

Clicking on one of the markers in the map view will limit the list to datasets at this specific location. 
Clicking on a dataset in the list will open a new window displaying detailed information about this dataset which includes
\begin{itemize}
    \item Geographic location
    \item Sensor modality
    \item Task / Application
    \item Size
    \item URL to access the data
    \item Number of views
    \item A graphical representation
    \item A brief description
\end{itemize}

A helpful function is the \emph{compare} option, which allows a direct side-to-side comparison of this information from two or more datasets in a newly opened window. As of the time of writing of this paper, there were 14 datasets in the catalogue. Designed as a community platform, EOD requires people creating or using remote sensing datasets to contribute actively by submitting new datasets to the database. 

\subsection{Dataset Submission}\label{sec:submission}
\begin{figure}[ht!]
    \centering
    \includegraphics[width=0.8\linewidth]{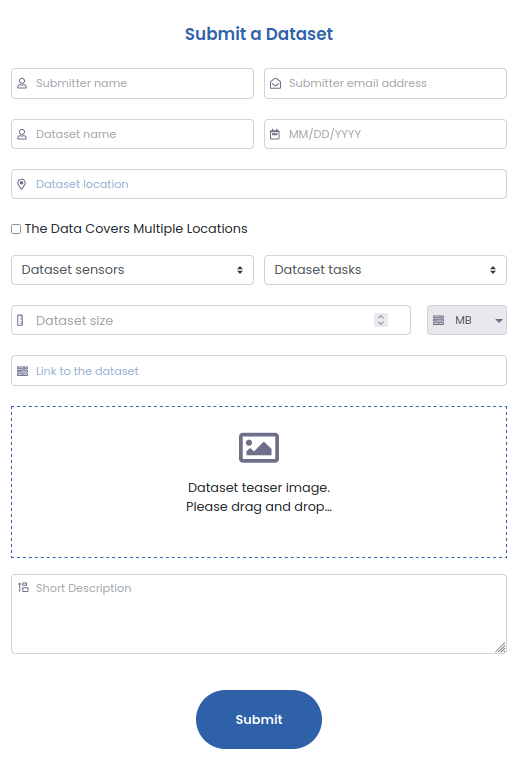}
    \caption{The EOD submission form.}
    \label{fig:EOD_Submission}
\end{figure}

As mentioned in the previous section, EOD is designed to be a collaborative community platform that can be used by every interested group or person to catalogue or advertise remote sensing datasets of any kind. New entries to the database can be realized via the submission form, which is depicted in Fig.~\ref{fig:EOD_Submission}.
As can be seen, several pieces of information are required for every dataset: 
\begin{itemize}
    \item \emph{Submitter name and email address:}\\
    This is just for reference for the database back end and administrators and will not be published. The email address will only be used, if, for example, important information about the dataset were missing or entered in an erroneous way.
    \item \emph{Dataset name:}\\
    Submitters should make sure to enter the correct name of the dataset, i.e. as the dataset authors would want the dataset to be known in the community.
    \item \emph{MM/DD/YYY:}\\
    This field shall indicate the date of publication of the dataset, \emph{not} the date of submission. This is important for use cases where possible changes might negatively impact the use of annotated data.
    \item\emph{Dataset location:}\\
    Here, the location (i.e. address) can be specified. Alternatively, if a dataset covers multiple locations, this can be specified as well.
    \item\emph{Dataset sensors and tasks:}\\
    In these fields, the sensor types that were used to produce the observations in the dataset as well as the tasks the data can be used for can be selected.
    \item\emph{Dataset size:}\\
    The size of the dataset can be provided either in terms of MB or GB. It should cover the size of the data to be downloaded.
    \item\emph{Link to dataset:}\\
    This is one of the most crucial pieces of information: Where can the user download the dataset for further usage.
    \item\emph{Dataset teaser image:}\\
    The submitter should always provide a small teaser image that grasps the nature of the dataset in a visual manner.
    \item\emph{Short description:}\\
    Here, the dataset should be described in just a few concise sentences. 
\end{itemize}
While it is very important that these information are entered with high care, datasets will not immediately appear in the public catalogue -- before approval, they will be checked by IADF members for correctness and consistency. 

\section{Call for Action}
As mentioned before, EOD is intended as a community platform and will thus thrive on contributions by the Earth observation and/or machine learning communities. Thus, this paper shall be concluded by a call to action: If you have created a dataset, or if you love to work with a dataset created by somebody else, please submit the dataset to EOD if it's not represented in the database yet! It should be our common goal to make EOD the most comprehensive database for EO-related datasets available in the world wide web.  

\section{ACKNOWLEDGMENTS}
The authors would like to thank the IADF co-chair Gemine Vivone and the IADF Working Group co-leads Seyed Ali Ahmadi, Francescopaolo Sica, and Xian Sun for helping to enter dataset information for the initial set before launching the webpage.

\bibliographystyle{IEEEbib}
\bibliography{refs}

\end{document}